\documentclass[5p,,preprint,12pt,onecolumn]{elsarticle}
\long\def\twocolumn[#1]{#1}

\makeatletter\if@twocolumn\PassOptionsToPackage{switch}{lineno}\else\fi\makeatother
\makeatletter\@twocolumnfalse \makeatother

\usepackage{tabulary,xcolor}
\usepackage{amsfonts,amsmath,amssymb}
\usepackage[T1]{fontenc}
\makeatletter
\let\save@ps@pprintTitle\ps@pprintTitle
\def\ps@pprintTitle{\save@ps@pprintTitle\gdef\@oddfoot{\footnotesize\itshape}}
\def\hlinewd#1{%
	\noalign{\ifnum0=`}\fi\hrule \@height #1%
	\futurelet\reserved@a\@xhline}

\AtBeginDocument{\ifNAT@numbers \biboptions{sort&compress}\fi}
\makeatother

\usepackage{ifluatex}
\ifluatex
\usepackage{fontspec}
\defaultfontfeatures{Ligatures=TeX}
\usepackage[]{unicode-math}
\unimathsetup{math-style=TeX}
\else 
\usepackage[utf8]{inputenc}
\fi 
\ifluatex\else\usepackage{stmaryrd}\fi

\usepackage{url,multirow,morefloats,floatflt,cancel,tfrupee}
\makeatletter

\AtBeginDocument{\@ifpackageloaded{textcomp}{}{\usepackage{textcomp}}}
\makeatother
\usepackage{colortbl}
\usepackage{xcolor}
\usepackage{pifont}
\usepackage[nointegrals]{wasysym}
\makeatletter

\def\mcWidth#1{\csname TY@F#1\endcsname+\tabcolsep}

\def\cAlignHack{\rightskip\@flushglue\leftskip\@flushglue\parindent\z@\parfillskip\z@skip}
\def\rAlignHack{\rightskip\z@skip\leftskip\@flushglue \parindent\z@\parfillskip\z@skip}

\usepackage{ifxetex}
\ifxetex\else\if@twocolumn\usepackage{dblfloatfix}\fi\fi

\AtBeginDocument{
	\expandafter\ifx\csname eqalign\endcsname\relax
	\def\eqalign#1{\null\vcenter{\def\\{\cr}\openup\jot\m@th
			\ialign{\strut$\displaystyle{##}$\hfil&$\displaystyle{{}##}$\hfil
				\crcr#1\crcr}}\,}
	\fi
}

\AtBeginDocument{%
	\@ifpackageloaded{endfloat}%
	{\renewcommand\efloat@iwrite[1]{\immediate\expandafter\protected@write\csname efloat@post#1\endcsname{}}}{}%
}%

\def\BreakURLText#1{\@tfor\brk@tempa:=#1\do{\brk@tempa\hskip0pt}}
\let\lt=<
\let\gt=>
\def\processVert{\ifmmode|\else\textbar\fi}

\@ifundefined{subparagraph}{
	\def\subparagraph{\@startsection{paragraph}{5}{2\parindent}{0ex plus 0.1ex minus 0.1ex}%
		{0ex}{\normalfont\small\itshape}}%
}{}

\newcommand\role[1]{\unskip}
\newcommand\aucollab[1]{\unskip}

\@ifundefined{tsGraphicsScaleX}{\gdef\tsGraphicsScaleX{1}}{}
\@ifundefined{tsGraphicsScaleY}{\gdef\tsGraphicsScaleY{.9}}{}
\def\checkGraphicsWidth{\ifdim\Gin@nat@width>\linewidth
	\tsGraphicsScaleX\linewidth\else\Gin@nat@width\fi}

\def\checkGraphicsHeight{\ifdim\Gin@nat@height>.9\textheight
	\tsGraphicsScaleY\textheight\else\Gin@nat@height\fi}

\def\fixFloatSize#1{}
\let\ts@includegraphics\includegraphics

\def\inlinegraphic[#1]#2{{\edef\@tempa{#1}\edef\baseline@shift{\ifx\@tempa\@empty0\else#1\fi}\edef\tempZ{\the\numexpr(\numexpr(\baseline@shift*\f@size/100))}\protect\raisebox{\tempZ pt}{\ts@includegraphics{#2}}}}

\AtBeginDocument{\def\includegraphics{\@ifnextchar[{\ts@includegraphics}{\ts@includegraphics[width=\checkGraphicsWidth,height=\checkGraphicsHeight,keepaspectratio]}}}

\def\URL#1#2{\@ifundefined{href}{#2}{\href{#1}{#2}}}

\def\UrlOrds{\do\*\do\-\do\~\do\'\do\"\do\-}%
\g@addto@macro{\UrlBreaks}{\UrlOrds}

\@ifundefined{quoteAttrib}
{}
{}

\@ifundefined{titlequoteAttrib}
{}{}

\newenvironment{title-quote}
{\list{}{\fontsize{10pt}{12pt}\selectfont\leftmargin.5in\itshape\rightmargin\leftmargin}%
	\item\relax}
{\endlist}

\makeatother

\usepackage{float}

\def\truncatedAt{1000}    
\makeatletter
\@ifpackageloaded{hyperref}{}{\usepackage[pdfborder={0 0 0},bookmarks=false]{hyperref}}
\def\putUpgradeInfoBox{\@ifundefined{truncatedAt}{\def\truncatedAt{1000}}{}
	\def\up@width@one{\if@twocolumn .95\columnwidth\else .8\columnwidth\fi}%
	\def\up@width@two{\if@twocolumn .5\columnwidth\else .3\columnwidth\fi}%
	\vskip 2pc\nopagebreak
	\noindent\centering\fboxsep 7pt \fbox{\parbox{\up@width@one}
		{\vskip 1.7pc
			\begin{center}
				\fontfamily{phv}\fontsize{18pt}{18pt}\selectfont%
				\resizebox{\up@width@one}{!}{\parbox{\up@width@one}
					{\centering{IT'S TIME TO UPGRADE}}}\\[2pc]
				\fontsize{9.5pt}{13pt}\selectfont%
				\linespread{1}%
				\def\baselinestretch{1}\selectfont%
				\resizebox{\up@width@one}{!}{\parbox{\dimexpr(\up@width@one)}
					{\centering Dear user, you are able to see document only till the word limit
						of \truncatedAt~words. To view entire document upgrade your~plan.}}
				\vskip 2.5pc
				{\href{https://typ.st/2noUNUS}{\includegraphics[width=\up@width@two]{upgrade-box-click.pdf}}}
				\vskip 1.5pc
				\rule{\dimexpr(\up@width@one-6pt)}{.5pt}
				\par
				\vspace*{-1pt}\par
				\fontsize{10pt}{13pt}\selectfont
				\fontfamily{phv}\selectfont\href{https://typ.st/2xUBN7Q}{www.typeset.io}
			\end{center}
}}}
\makeatother
    
\begin{document}
	
	\begin{frontmatter}

		\title{Cross-layer scheme for low latency multiple description video streaming over Vehicular Ad-hoc NETworks (VANETs)}

		\author[affe797ec712bf89cd756f86529dde6ef4f,aff56a7721a4930ecb26ce66fb37a3ca85a]{Mohamed Aymen Labiod}
		\ead{MohamedAymen.Labiod@uphf.fr}
		\author[affe797ec712bf89cd756f86529dde6ef4f]{Mohamed Gharbi}
		\author[affe797ec712bf89cd756f86529dde6ef4f]{Fran\c{c}ois-Xavier Coudoux}
		\author[affe797ec712bf89cd756f86529dde6ef4f]{Patrick Corlay}
		\author[aff56a7721a4930ecb26ce66fb37a3ca85a]{Noureddine Doghmane}
		
		\address[affe797ec712bf89cd756f86529dde6ef4f]{Univ. Polytechnique Hauts-de-France, CNRS, Univ. Lille, YNCREA, Centrale Lille, UMR 8520 - IEMN, DOAE, F-59313 Valenciennes, France}
		
		\address[aff56a7721a4930ecb26ce66fb37a3ca85a]{Automatic and signals laboratory Annaba (LASA), Department of electronic, Badji Mokhtar University Annaba, Algeria.}
		
		\begin{abstract}
			
There is nowadays a growing demand in vehicular communications for real-time applications requiring video assistance. The new state-of-the-art high-efficiency video coding (HEVC) standard is very promising for real-time video streaming. It offers high coding efficiency, as well as dedicated low delay coding structures. Among these, the all intra (AI) coding structure guarantees minimal coding time at the expense of higher video bitrates, which therefore penalizes transmission performances. In this work, we propose an original cross-layer system in order to enhance received video quality in vehicular communications. The system is low complex and relies on a multiple description coding (MDC) approach. It is based on an adaptive mapping mechanism applied at the IEEE 802.11p standard medium access control (MAC) layer. Simulation results in a realistic vehicular environment demonstrate that for low delay video communications, the proposed method provides significant video quality improvements on the receiver side.
			
			
		\end{abstract}
		

	\end{frontmatter}
	
	\section{ Introduction}
	
The development of vehicular communications leads to increasingly high expectations for efficient video vehicle-to-everything (V2X) transmissions with high-level quality of service (QoS) guarantees \unskip~\cite{cunha_data_2016}. The challenge is even more difficult for applications requiring low latency \unskip~\cite{quadros_qoe-driven_2016}. Such applications concern traffic management, security, overtaking maneuver or entertainment applications. Also, the emergence of autonomous vehicles leads to the important need of efficient vehicle remote control strategies. Moreover, it is mandatory that a human operator can remotely regain the control of the autonomous vehicle in some critical cases like bad weather conditions, road accident, traffic jams, roadworks, or passenger support. For these purposes, the latency should not exceed 100 ms, which represents the average response time of a human observer \unskip~\cite{eze_advances_2016, parvez_survey_2018}. This also requires video delivery with good quality and high frame rate. So as to ensure a better interpretation of the context and minimize the reaction time. While, the vehicular ad-hoc network (VANET) has a limited reliability and low bandwidth. High packet loss rates are also possible \unskip~\cite{junior_game_2018}. In the particular case of video transmission, there are generally three kinds of approaches to reduce packet loss effects: 

\begin{itemize}
	\item At the transport layer protocol based on the reliable automatic repetition request (ARQ) such as transmission control protocol (TCP). However, this option has the disadvantage of adding a significant delay which is not acceptable for low latency applications. 
	
	\item By adding a forward error correction (FEC) that quantitatively increases the throughput bitrate. 
	
	\item With error resilient coding (ERC), which is used as source-level protection and can take many forms, such as multiple description coding (MDC).
	
\end{itemize} 

Indeed, MDC is an effective approach that aims to improve the quality of multimedia streaming based on path diversity \unskip~\cite{Salkuyeh}; i.e. when multiple paths are available between the sender and the receiver. The principle of MDC is to separate the source signal into several representations called descriptions such that:

\begin{itemize}
	\item Each description can be decoded independently when received;
	\item The higher the number of received descriptions, the better the reconstructed video quality.
\end{itemize}

The choice of the video coding solution is just as important. Among the existing video coding standards, the latest HEVC has demonstrated significant compression gains and high resiliency for a high loss rate transmission \unskip~\cite{Sullivan, Psannis}. On the other hand, several technologies exist to allow V2X transmission, such as: IEEE 802.11p, long term evolution (LTE) or the future 5G and cellular V2X (C-V2X). Currently, the IEEE 802.11p seems to be the technology that has the most advantage by the fact that it is already deployed and can be generalized \unskip~\cite{campolo_todays_2015}. The fact that it is free to use is also an important advantage in comparison to cellular technologies. Moreover, the IEEE 802.11p standard provides better delay performance and allows end-to-end delay of less than 100 ms \unskip~\cite{HameedMir2014, 6170893, 6155707}. In this work, a cross-layer approach is proposed based on multiple description coding and an adaptive algorithm at the IEEE 802.11p MAC layer. The low complexity of the proposed scheme relies on the different choices adopted, whether for the coding structure, the MDC system or the simplicity of the mapping algorithm. This makes it possible to overcome the low latency constraint. Furthermore, the proposed algorithm is based on a differentiation between the descriptions and the channel state being determined by the MAC layer buffer queue length. Simulation results in a realistic vehicular environment show that the average PSNR gains can reach 7.75 dB compared to conventional video transmission schemes.

The rest of the paper is organized as follows. In Section 2 we give a brief overview of the related work existing in the literatures on multimedia transmission over vehicular networks. In Section 3 we give a detailed description of the proposed solution. It is based on an adaptive cross-layer scheme which relies on two parts of the transmission chain. A multi-description encoder based on the HEVC standard is used at the application layer in order to make the video bitstream more robust against channel losses. Then, a modification of the video mapping algorithm between the different access category (AC) is applied at the MAC layer level. The combination of both offers better performances and improves QoS guarantees from an end-to-end point of view. The parameters setting is presented in Section 4. The simulation results are presented in Section 5. Finally, we give our conclusion in Section 6.
	
	\section{Related work }
	
The improvement of multimedia transmissions over vehicular networks has been widely discussed in the literature \unskip~\cite{Oche, 7029128, 271779:6070781, Vinel}. The authors in \unskip~\cite{Oche} proposed a QoS analysis evaluation of internet protocol television (IPTV) transmission in VANET. In \unskip~\cite{7029128} an error recovery mechanism called multichannel error recovery video streaming (MERVS) have been proposed for real-time video streaming in VANET. The proposed solution differentiates between frames and transmits I-frames using the TCP and the remainder of the frame types using the user datagram protocol (UDP). Zaidi et al. \unskip~\cite{271779:6070781} proposed another error recovery protocol where also a distinction between images is used, based on a retransmission technique. While Vinel et al. \unskip~\cite{Vinel} proposed an end-to-end model for overtaking maneuver assistance applications. The model is based on real time codec adaptation to the channel variation in the IEEE 802.11p standard.

The MDC is one of the main robust error-resilient encoding methods that can be used in image \unskip~\cite{BERROUCHE2014976} and video error protection \unskip~\cite{259576:5809267}. Particularly in video applications requiring low latency where retransmission is not possible \unskip~\cite{259576:5809268}. This solution has proven itself in various networks with severe conditions and a high packet loss rate as shown by Kazemi et al. \unskip~\cite{Kazemi2014}. Their paper also presents the different existing MDC schemes. One of the alternatives to MDC is scalability with the H.264 extension named scalable video coding (SVC) or the HEVC extension named scalable HEVC (SHVC) \unskip~\cite{Chiang}. SVC or SHVC decomposes the video signal into several layers: a so-called base layer and one or several additional enhancement layers. However, in this case, it is mandatory to receive the base layer in order to decode a video signal of minimal quality.

\mbox{}\protect\newline In some works, there is a recent trend to use MDC in mobile video streaming applications \unskip~\cite{259576:5809270,259576:5809271}. As for Qadri et al. \unskip~\cite{5508799}, they consider the MDC in multi-source video streaming through VANET. For this, they used spatial decomposition with the checkerboard’s flexible macroblock ordering (FMO), an H.264/AVC error-resilience technique.  In \unskip~\cite{fatani_robust_2012, fatani_multiple_2011}, the authors combine MDC based on H.264/AVC with multiple input multiple output (MIMO) transmission using several antennas to enhance train to wayside video transmissions in tunnels. Zhoua et al. \unskip~\cite{ZHOU2009638} proposed a cross-layer technique for a wavelet coded multistream video transmission over multihop wireless networks. The proposed solution combines a cross-layer distributed flow control algorithm at the MAC and transport layers as well as a rate-distortion joint source-channel coding approach build on several independent streams and a forward error correction (FEC).	

On the other hand, the survey \unskip~\cite{259576:5809272} describes recent work on cross-layer communication design for VANETs, some improvements have been made to MAC protocols for security applications in VANET. Gupta et al. \unskip~\cite{259576:5809273} presented a comprehensive study of some of these works. Furthermore, the IEEE 802.11p extension for vehicular networks with variable QoS supports differentiated service classes at the MAC layer \unskip~\cite{259576:5809266}. Labiod et al. \unskip~\cite{LABIOD201928} proposed an cross-layer mechanism to improve HEVC video streaming in VANETs with a low delay constraint. The algorithm adapts to the new temporal prediction structures introduced in the HEVC. The mechanism proposed takes into account both the importance of the frame in a video according to the temporal prediction structures and the state of the channel determined by the length of the queue of the MAC layer. The authors admitted that the AI structure was the most appropriate for low latency applications. However, no classification has been adopted for this temporal prediction structure.

Improvements have also been made to enhanced distributed channel access (EDCA) in the case of 802.11e video transmission previously. Indeed, Lin et al. \unskip~\cite{259576:5809274} have proposed an adaptive cross-layer-mapping algorithm to improve the video streaming quality of MPEG-4 Part 2 video over IEEE 802.11e wireless networks. However, they do not take into account either a multiple description encoding or a latency constraint.
\mbox{}\protect\newline Before them Ksentini et al. \unskip~\cite{259576:5809276}, were the first ones to have the idea to use other available access categories (ACs). Indeed, the authors proposed a cross-layer static architecture based on an H.264 video stream. Bernardini et al. \unskip~\cite{259576:5809277} suggested the same approach with an MDC scheme.  Whereas, Miliani et al. \unskip~\cite{259576:5809275} established a classification strategy based on the percentage of null quantized DCT coefficients for MDC encoded packets taking advantage of the QoS differentiation provided by the IEEE 802.11e standard. Nevertheless, the various works listed do not take into account the state of the channel. Moreover, again, no latency constraint is taken into account and the vehicular IEEE 802.11p standard is not considered.

\section{Description of the proposed solution}
In this section, we first briefly describe the technologies used in the proposed solution. Then, we present in detail the proposed adaptive cross-layer MDC-based approach. We also address a static algorithm that will serve as a comparative.

\subsection{The IEEE 802.11p standard}

Several solutions have been proposed for wireless access in VANETs. However, IEEE established a set of standards for vehicle-to-vehicle (V2V) and vehicle-to-infrastructure (V2I) communication called WAVE (Wireless Access in the Vehicular Environment). WAVE is based on two standard categories: 802.11p for PHY and MAC layers and IEEE 1609 for network management, security as well as other side of VANETs. The ITS-G5, its equivalent in Europe, shows some difference in the upper layers \unskip~\cite{festag_standards_2015}.

Based on dedicated short-range communication (DSRC) standard in the 5.850-5.925 GHz frequency range, the spectrum band reserved for IEEE 802.11p is divided into seven 10 MHz channels, numbered between 172 and 184 \unskip~\cite{259576:5809278}. Except the 178 channel, named control channel (CCH), dedicated to control information and providing access to critical safety applications. The other channels can be used for different services and are allocated for data transmission named service channels (SCH). The physical layer is based on the 802.11a standard and uses OFDM (Orthogonal Frequency-Division Multiplexing) modulation with however a change of bandwidth from 20 MHz to 10 MHz. The purpose being to reduce the spreading delay in VANETs which also results in reducing the interference between adjacent channels. The standard provides communication at theoretical distances of up to 1000 meters with a bitrate variation of 3 Mbps to 27 Mbps. As for the MAC layer, it is based on the 802.11e standard and uses EDCA for packet transmission. EDCA is an improvement of distributed channel access (DCA) to provide sufficient QoS. However, instead of single queue storing data frame, EDCA has four queues representing different levels of priority named AC. Each of these ACs is dedicated to a kind of traffic, namely Background (BK or \textit{AC}[0]), Best Effort (BE or \textit{AC}[1]), Video (VI or \textit{AC}[2]) and Voice (VO or \textit{AC}[3]). An illustration of the IEEE 802.11p four access categories is given in Figure~\ref{figure-MAC-layer}. Voice traffic is given the highest priority while is the Background lowest one. Different arbitration inter-frame space number (AIFSN) and contention window (CW) values are selected for different types of ACs. The values of each AC are shown in Table~\ref{table-wrap-e4d6724dea32374aa7fa31c0b61a4a2b} \unskip~\cite{259576:5809273}. We verify that the video AC has a lower \textit{T\ensuremath{_{\rm AIFS}}} than the BE and BK, which means less waiting for access to the medium, this leads to a higher priority. We can also see that the different values AIFSN and CW are different between the CCH and the SCH.

\bgroup
\fixFloatSize{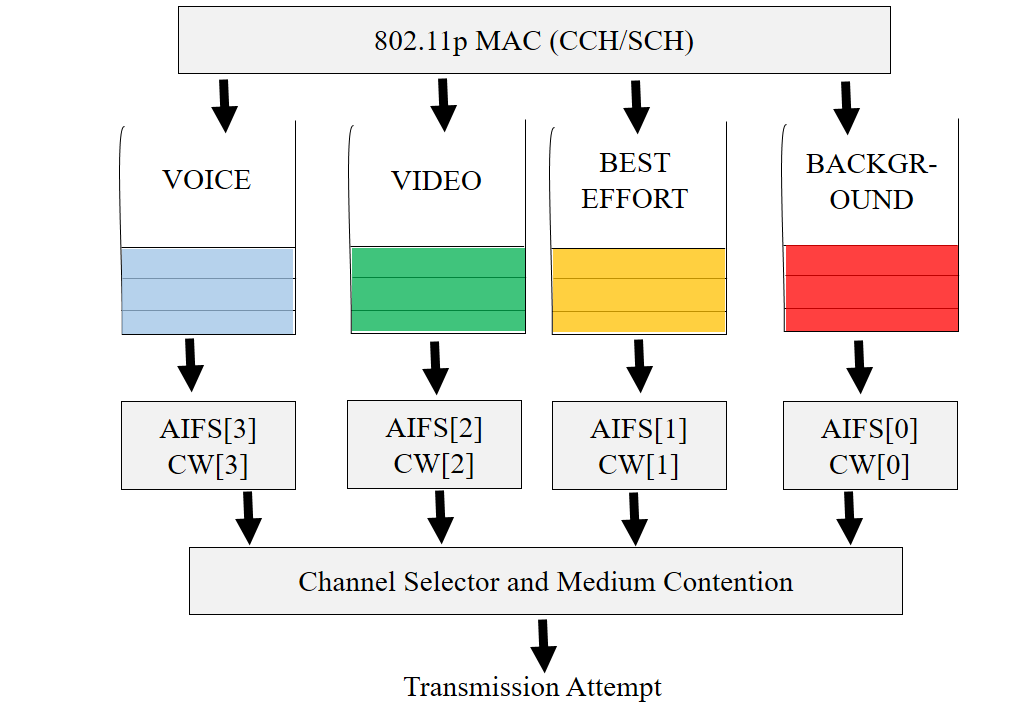}
\begin{figure*}[!htbp]
	\centering \makeatletter\IfFileExists{images/Figure1.png}{\includegraphics[width=0.51\linewidth]{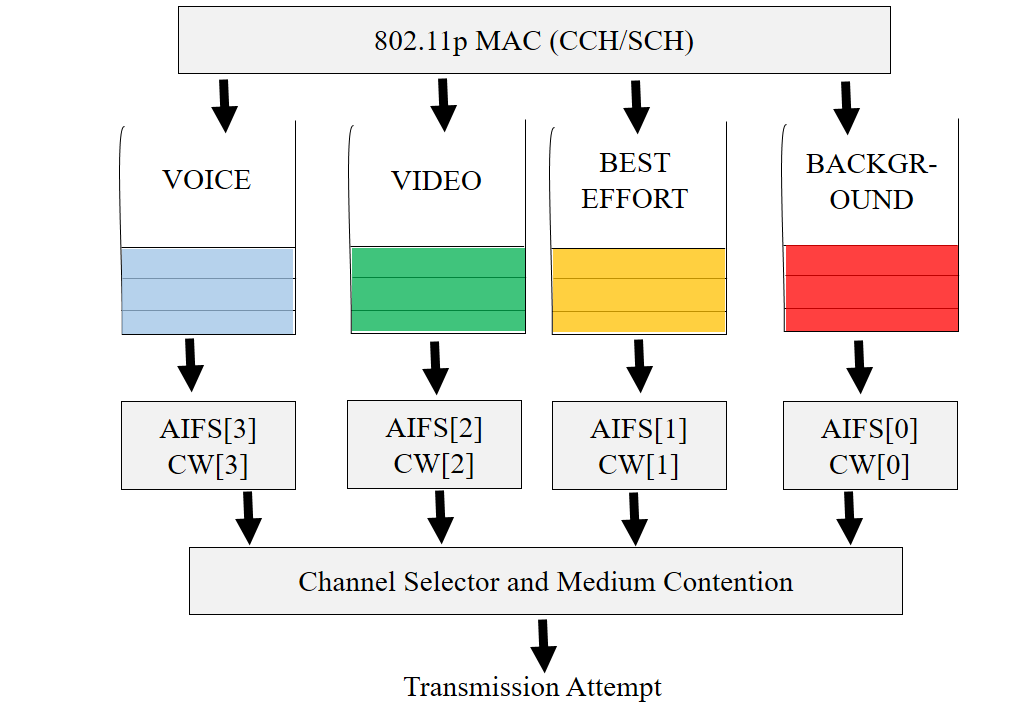}}{}
	\makeatother 
	\caption{{Four access categories in IEEE 802.11p.}}
	\label{figure-MAC-layer}
\end{figure*}
\egroup

\begin{table*}[!htbp]
	\centering
	\caption{{ IEEE 802.11p Access Categories. } }
	\label{table-wrap-e4d6724dea32374aa7fa31c0b61a4a2b}
	\begin{tabular}{llllllllll} 
		\hline
		\multirow{2}{*}{\begin{tabular}[c]{@{}c@{}}AC \\ Number\end{tabular}} & \multirow{2}{*}{\begin{tabular}[c]{@{}c@{}}Access\\ Category\end{tabular}} & \multicolumn{4}{c}{CCH}            & \multicolumn{4}{c}{SCH}            \\
		&                                                            & CWmin & CWmax & AIFSN & {\begin{tabular}[c]{@{}c@{}}{T\ensuremath{_{\rm AIFS}}} \\ (µs)\end{tabular}}& CWmin & CWmax & AIFSN & {\begin{tabular}[c]{@{}c@{}}{T\ensuremath{_{\rm AIFS}}}  \\ (µs)\end{tabular}} \\	\hline
		0                                                                       & \begin{tabular}[c]{@{}l@{}}Background\\ Traffic (BK) \end{tabular}          & 15    & 1023  & 9     & 149                                              & 31    & 1023  & 7     & 123                                                \\
		1                                                                       & \begin{tabular}[c]{@{}l@{}}Best Effort \\ (BE) \end{tabular}                & 7     & 15    & 6     & 110                                              & 31    & 1023  & 3     & 71                                                 \\
		2                                                                       & Video(VI)                                                                   & 3     & 7     & 3     & 71                                               & 15    & 31    & 2     & 58                                                 \\
		3                                                                       & Voice(VO)                                                                   & 3     & 7     & 2     & 58                                               & 7     & 15    & 2     & 58  
		\\
		\hline                                              
	\end{tabular}
\end{table*}

\subsection{Video Encoding}

\subsubsection{MDC scheme}

Since the MDC is one of the relevant solutions for  streaming video delivery over lossy networks it has a wide applications in the case of packet-switching networks. However, because of fading channels, wireless communication is usually affected by network congestion, available bandwidth, backbone network capacity and route selecting, which results in data packet losses. The MDC helps ensure robust transmission over hostile channels and provides a low-complexity solution that alleviates visual artifacts \unskip~\cite{noauthor_scalable_2009}. The goal of MDC is to create several independent descriptions that: exploit the intrinsic characteristics of a video signal temporal or spatial resolution, frequency content, signal-to-noise ratio. The different descriptions can be of equal importance (balanced MDC schemes) or not (unbalanced MDC schemes) \unskip~\cite{wang_multiple_2005}. MDC addresses the problem of encoding a source for transmission over a multiple channels communication system. In the most common implementation, two descriptions are generated and both have equivalent bitrate and visual importance so that each description alone provides low but acceptable quality and both descriptions together lead to higher quality. The packetization and sending of the two descriptions is done individually via the same or separate physical channels. MDC schemes can be classified as follows: odd/even fields separation (also called spatial splitting), odd/even frames separation (also called temporal splitting), redundant slices, spatial MDC and scalar quantizer MDC. Due to their simplicity in producing multiple streams, odd-even frames are particularly appealing for real-time interactive applications. Basically, the odd–even-frame based MDC scheme divides video sequence frames into two descriptions. The first description contains the even frames of the sequence and the second the odd ones. This scheme makes it easier to avoid any mismatch at the expense of reduced coding efficiency. Indeed, compared to standard video encoders the temporal prediction distance is more significant which results in a larger video bitrate \unskip~\cite{bai_distributed_2011}.

\subsubsection{HEVC overview}

The recent HEVC standard provides about 50\% bitrate savings, compared with its predecessor H.264 for the same image quality. HEVC, like H.264, presents a two-layer high-level design in the form of a hybrid video coding system. This consists of a video coding layer (VCL) and a network abstraction layer (NAL). The compressed payload data is thus encapsulated in network adaptation layer units (NALU). Moreover, HEVC made it possible to envisage the transmission of video and particularly real-time video in circumstances presenting severe transmission conditions, such as low bandwidth networks or with a high packet loss rate. Indeed, a comparison of HEVC with various standard: H.264/AVC, MPEG-4 part 2 and H.263 coding standards have allowed Psannis et al. \unskip~\cite{Psannis} to prove that HEVC outperforms its predecessors in terms of error propagation resilience in a wireless environment. The authors \unskip~\cite{Paredes} studied the received quality from compressed video data streams after transmission over VANETs. The video encoders H264, H265, and VP9 were considered. Hence, their research shows that new generation H.265 and VP9 encoders offer better video reception quality. Torres et al. in \unskip~\cite{Torres} reached the same conclusion after a comparison between the H.264 and H.265 video coding standards. Pinol et al. \unskip~\cite{pinol} evaluated video coding standard HEVC for video streaming in VANET based on the packet loss problem. Several encoding factors have also been studied.

Furthermore, the HEVC has defined three predictive structures depending on the intended application, in terms of efficiency, computational complexity, processing time or error resilience techniques. The three types of encoding structure are \unskip~\cite{259576:5809282}:

\begin{itemize}
	\item Random access (RA): presents the highest compression gain at the expense of an associated encoding delay.
	\item Low delay (LD): this configuration has no delay due to the coding structure i.e. with the same coding and decoding order, but has a lower resilience error compared to the other two structures. This is explained by the inter-frame dependence.
	\item All intra (AI): in this configuration, all pictures are coded independently with intra prediction only. Thus, it has a lowest complexity and it is suitable for low delay applications at the expense of a higher rate compared to the two previous structures. 
\end{itemize}

Given its low complexity, high error resilience and low encoding time the AI structure was chosen in this work.

\subsection{Description of the proposed system}

In the proposed low complexity adaptive system, the odd-even frames MDC is used in VANET. As illustrated in Figure~\ref{figure-1eb5a4b12034e1ad327e14a8f75ce28d}, the adopted cross-layer scheme exploits the two descriptions video packets stream in the mapping algorithm at the IEEE 802.11p MAC layer. At the application layer, the odd and even frames are separated into two descriptions and sent to two independent AI H.265/HEVC encoders. Each encoder generates a packet stream that will be sent to the lower levels of the protocol stack. 

\bgroup
\fixFloatSize{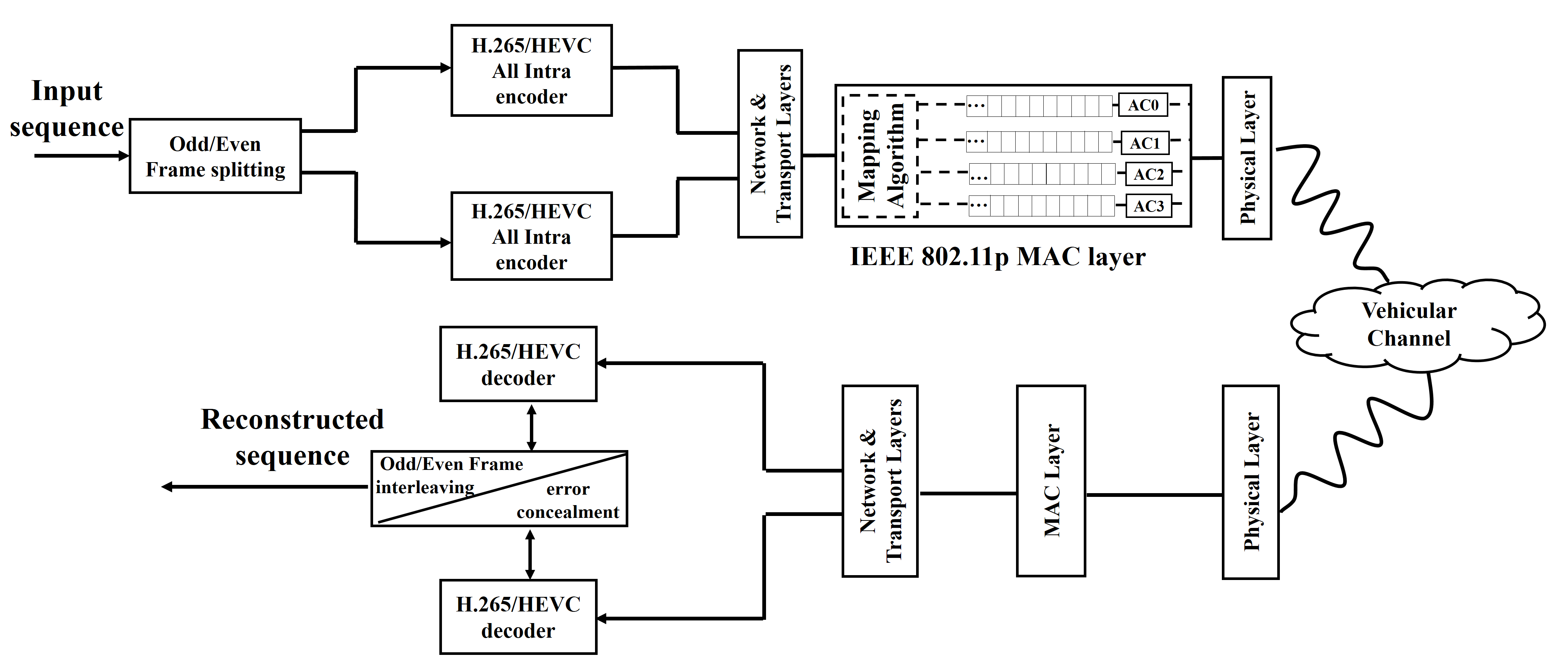}
\begin{figure*}[!htbp]
	\centering 
	\makeatletter\IfFileExists{images/2ded5fbc-8c46-49ca-9913-252f3ef33f60-uproposed-system-letter-0.eps}{\includegraphics[width=0.99\linewidth]{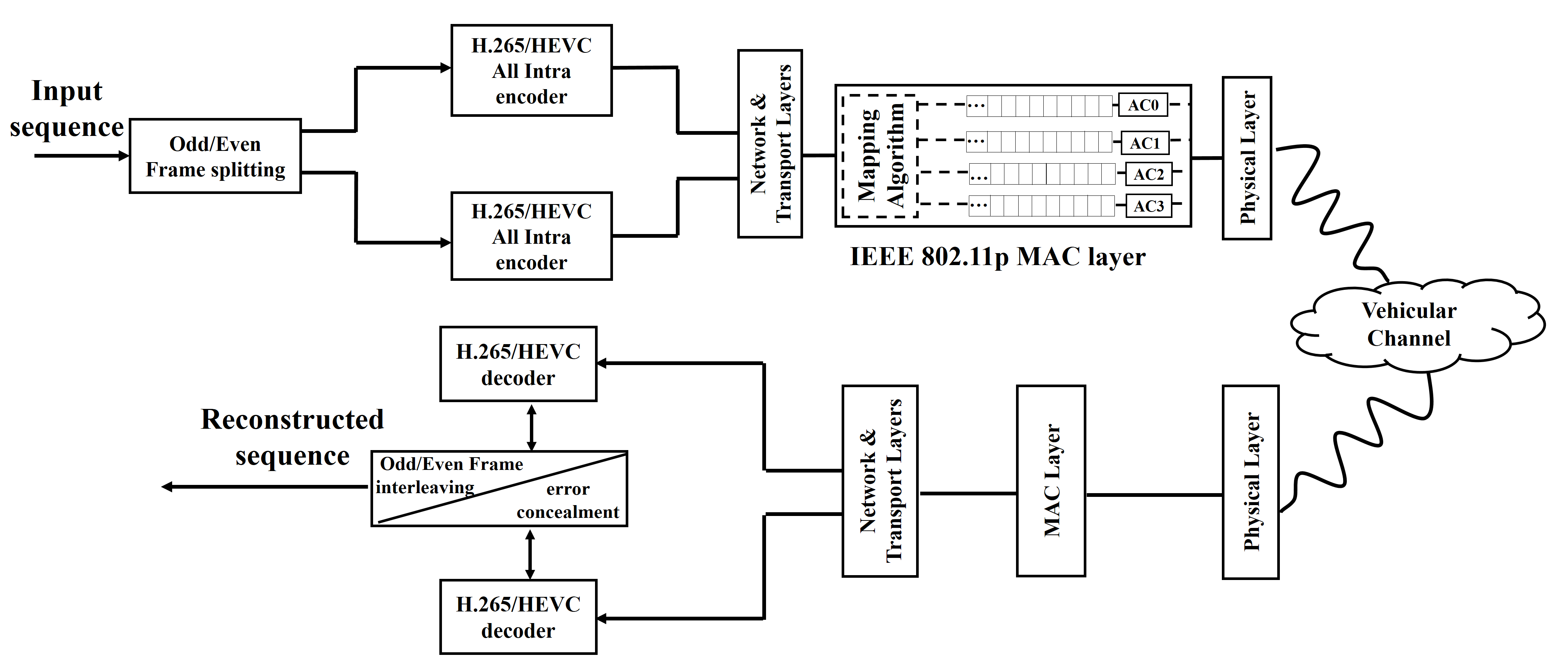}}{}
	\makeatother 
	\caption{{Representation of the adopted transmission MDC scheme.}}
	\label{figure-1eb5a4b12034e1ad327e14a8f75ce28d}
\end{figure*}
\egroup

At the receiver, if both H.265/HEVC decoders receive all the packets correctly, both descriptions can be correctly decoded and the coded sequence can be reconstructed without any additional loss of quality. In the case where some parts of a description are missing, the description is suppressed  and an error concealment mechanism occurs by applying a copy frame from the image of the other description as illustrated in Figure~\ref{figure-5650f277e75d8c94f5f42e6e1e8331c4}. This mechanism allows us to have a reconstructed sequence less degraded. Indeed, high frame rate encoding makes it possible, in the case of poor quality channel, to recover a good quality sequence with a low frame rate. Since we use the frame copy the worst case being to have half initial frame rate, this ensures minimal video quality.

\bgroup
\fixFloatSize{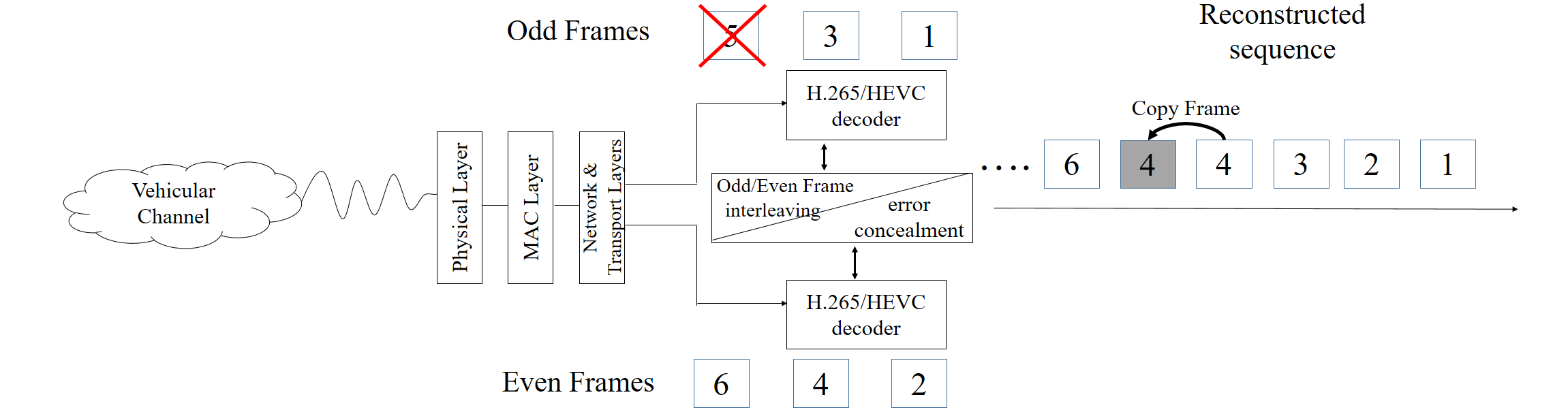}
\begin{figure}[!htbp]
	\centering 
	\makeatletter\IfFileExists{images/Proposed-system-letter2.png}{\includegraphics[width=1.\linewidth]{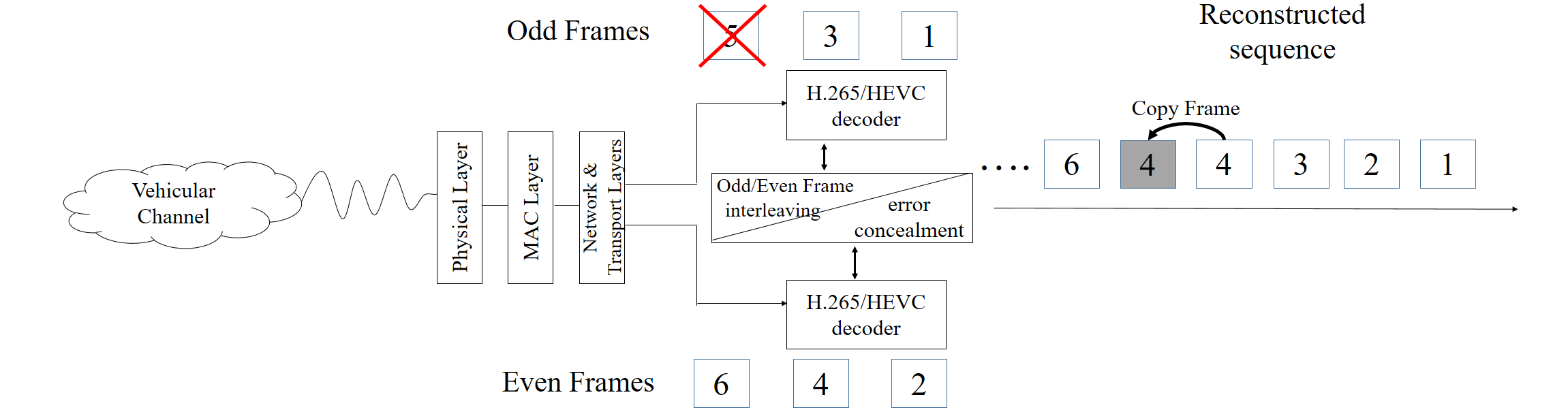}}{}
	\makeatother 
	\caption{{The adopted error concealment mechanism.}}
	\label{figure-5650f277e75d8c94f5f42e6e1e8331c4}
\end{figure}
\egroup

The proposed algorithm applies a differentiated processing to both descriptions. The goal is to protect, at least, one description to the detriment of the other. To do this, we propose to use the others ACs in addition to the one used by the video namely the \textit{AC}[2]. The other ACs we propose to use are of less priority, to know: \textit{AC}[1] and \textit{AC}[0] respectively for the best effort flow and the background traffic. 

\subsubsection{Static mapping algorithm}

In the static mapping algorithm used for comparison, we route packets of the odd description on the \textit{AC}[2] and packets of the even description on the \textit{AC}[1]. This increases the chances of correctly receiving the odd description. The principle is inspired by the work of Ksentini et al. \unskip~\cite{259576:5809276}. Figure~\ref{figure-stat-prop} illustrated the static mapping algorithm.

\bgroup
\fixFloatSize{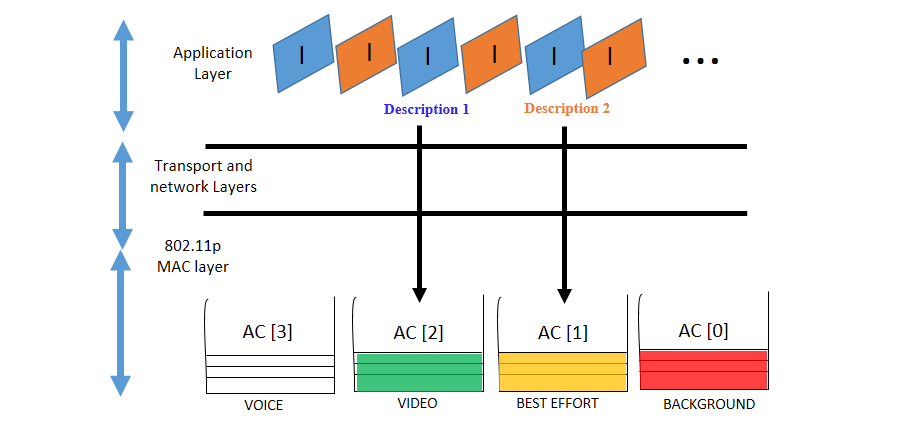}
\begin{figure}[!htbp]
	\centering \makeatletter\IfFileExists{images/Static-prop.png}{\includegraphics[width=.65\linewidth]{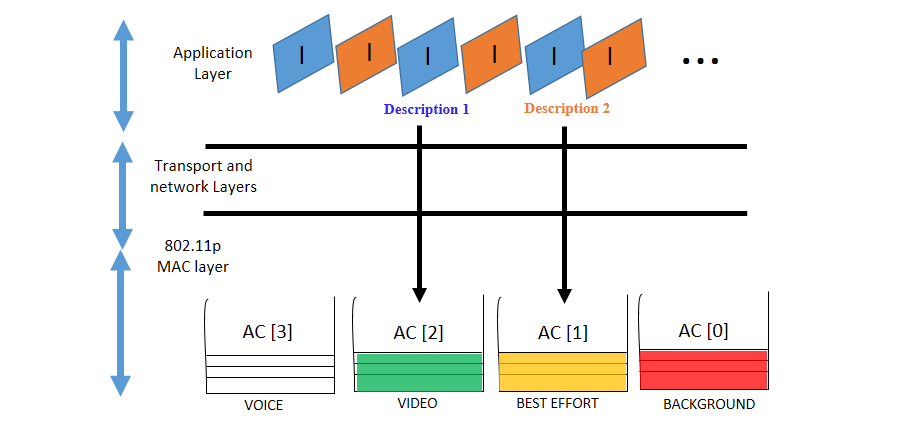}}{}
	\makeatother 
	\caption{{Illustration of the static cross-layer algorithm.}}
	\label{figure-stat-prop}
\end{figure}
\egroup

\subsubsection{Proposed mapping algorithm}

For the proposed mapping algorithm, it dynamically allocates each video packet to the most appropriated AC at the MAC layer. It takes into account the state of the network traffic load and the description of each frame packet. In addition, to differentiate between the two descriptions, each description has a different probability of mapping to lower priority ACs, defined as ${P_{Descrip}}$. The probability is between 0 and 1 and depends on the importance of the frame whereas ${P_{Even}}>{P_{Odd}}$.

In addition, as previously mentioned, the mapping takes into consideration the state of the channel thanks to the state of filling of the AC queues. Indeed, the more the MAC queue is filled, the more the network is overloaded. In order to deal with network congestion, we have established two thresholds, denoted threshold\_high and threshold\_low. Based on the principle of the random early detection (RED) mechanism \unskip~\cite{251892} and inspired by Lin et al. \unskip~\cite{259576:5809274} algorithm, the proposed adaptive mapping algorithm is based on the following expression:
\begin{equation}
{P_{new}} = {P_{Descrip}} \times \frac{{qlen\left( {AC\left[ {2} \right]} \right) - qt{h_{low}}}}{{qt{h_{high}} - qt{h_{low}}}}\
\label{moneq2}
\end{equation}
where ${P_{Descrip}}$ is the initial probability of each description, ${P_{new}}$ is the updated probability,  ${qlen\left( {AC\left[ {2} \right]} \right)}$ is the actual state of the video queue length, $qt{h_{high}}$ and $qt{h_{low}}$ are arbitrarily chosen thresholds that define the manner and degree of mapping to ACs with priority lower. Figure~\ref{figure-adap-prop} illustrated the proposed mapping algorithm.

\bgroup
\fixFloatSize{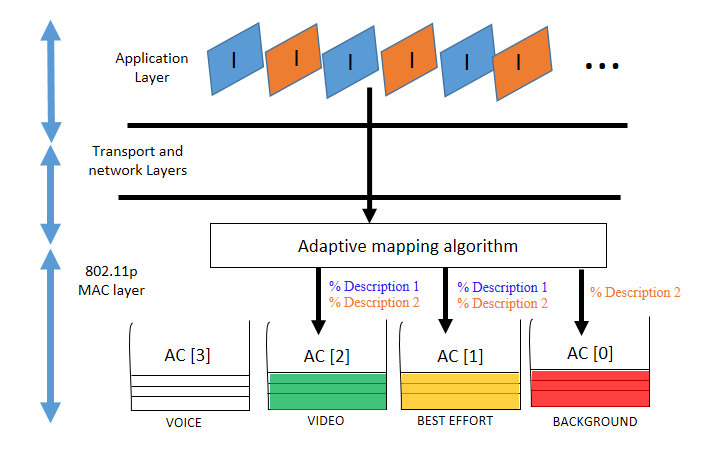}
\begin{figure}[!htbp]
	\centering \makeatletter\IfFileExists{images/adaptive-prop.png}{\includegraphics[width=.65\linewidth]{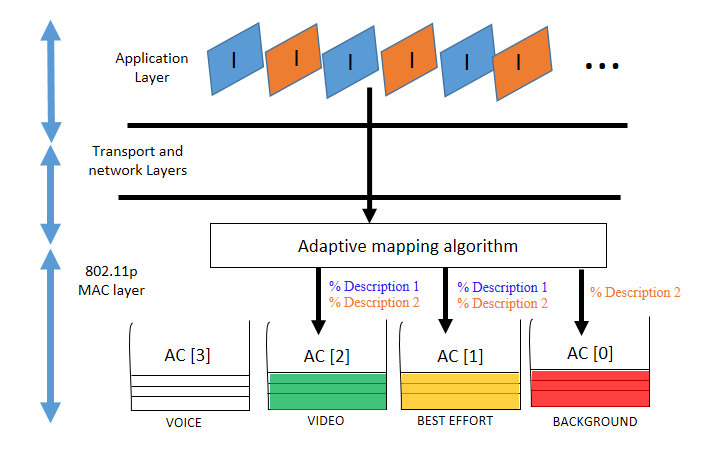}}{}
	\makeatother 
	\caption{{Illustration of the proposed cross-layer algorithm.}}
	\label{figure-adap-prop}
\end{figure}
\egroup

To explain the behavior of the algorithm, when ${qlen\left( {AC\left[ {2} \right]} \right)}$ is less than $qt{h_{low}}$, all packets are mapped in \textit{AC}[2]. When ${qlen\left( {AC\left[ {2} \right]} \right)}$ is between $qt{h_{high}}$ and $qt{h_{low}}$, ${P_{new}}$ sets the packet probability mapped to \textit{AC}[1]. And finally, when ${qlen\left( {AC\left[ {2} \right]} \right)}$ is greater than $qt{h_{high}}$, video packets are mapped in \textit{AC}[1] with ${P_{Descrip}}$ probability of being mapped to \textit{AC}[0]. Figure~\ref{figure-1eb5a4b12034e1ad327e14a8f75ce28d} illustrates the entire proposed system.

\section{Simulation and parameters setting }
In order to model a realistic vehicle environment, we have implemented a realistic simulation of a video transmission in a vehicular Framework. Our Framework is composed of three main blocks, which are: a vehicular traffic simulator, a network simulator and a video encoder/decoder. The SUMO (Simulation of Urban Mobility) environment has been considered \unskip~\cite{259576:5809280}. This open-source simulator models the behavior of vehicles with urban mobility and takes into account the interaction of vehicles with each other, junctions, traffic signals, etc. Therefore, we use realistic urban traffic maps of Valenciennes town (France) imported from OpenStreetMap (OSM) \unskip~\cite{259576:5818119}. We used the road mobility generated by SUMO in our network simulator. As a network simulator, we decided to use NS2 to which we integrated the Evalvid tool \unskip~\cite{259576:5809281} that allowed us to simulate the video transmission and reconstruct it at the receiver side. The videos meanwhile were encoded with the reference HEVC model (HM) 16.16\unskip~\cite{259576:5818121}. 

For the choice of the appropriate sequences, we choose class C sequences according to the classification introduced by the video coding collaborative team (JCT-VC)\unskip~\cite{259576:5809282}. This video sequence class is intended to evaluate the performance of defined mobile video applications with an image resolution of 832x480 pixels. During the different simulations, we made the choice to use the following three sequences: PartyScene, BasketballDrill, and BQMall, which have a fairly important frame rate of respectively 50, 50 and 60 fps. The radio propagation model used is TwoRayGround allowing to give a realistic representation of free space vehicular channel \unskip~\cite{Sommer}, the used standard is the IEEE 802.11p. As for the transport layer protocol and in order to guarantee a minimum latency, we have chosen to work with the UDP. Also, the Ad Hoc On-Demand Distance Vector (AODV) routing protocol has been chosen based on Zaimi et al.\unskip~\cite{Zaimi} work. Moreover, it is one of the most widely used and most researched protocols in VANET \unskip~\cite{DARABKH2018277}.
\mbox{}\protect\newline In order to evaluate the reconstructed video quality at the reception, we use two video quality evaluation metrics: peak signal to noise ratio (PSNR) and structural similarity (SSIM). On the other hand, \unskip~\cite{Joseph, yim_evaluation_2011} have shown that it is also important to study the temporal variability of the video quality. Indeed, an average quality alone cannot reflect a realistic assessment of the quality of experience (QoE). Furthermore, \unskip~\cite{yim_evaluation_2011} shows that significant temporal variability of video quality may have a more negative effect on QoE than constant and lower average video quality. In order to evaluate the variability of video quality,  the standard deviation of the MSE was used as a criterion.

\begin{itemize}
	\item PSNR is the most widely used and representative metric and calculated usually frame by frame between the reconstructed received video and the original one. For the $k^{th}$ frame, $k=0,1,\cdots,K-1$, it is defined as follows:
\end{itemize}
\begin{equation}
{PSNR_{k}} = 10 \times{\log _{10}}\left( {\frac{{{d^2}}}{{MSE_{k}}}} \right)\
\label{moneq7}
\end{equation}
where $ {d} $ is the maximum possible value of a pixel. In the standard case of an image, the components of a pixel are encoded on 8 bits, in this case $ {d=255} $. The mean squared error (MSE) of two frames ${I_r}$ and ${I_o}$ of size $M\times N$, i.e. the frame MSE, is defined as follows:
\begin{equation}
{MSE_{k}} = \frac{1}{{M\times N}}\sum\limits_{m = 0}^{M - 1} {\sum\limits_{n = 0}^{N - 1} {({I_o}(} } m,n,k) - {I_r}(m,n,k){)^2}\
\label{moneq8}
\end{equation}
where integer indices $m$ and $n$ are the column and row indices.

For our part, we also calculate the average video PSNR between the reconstructed received video and the original video. As demonstrated by \unskip~\cite{Nasrabadi} for channel error or packet loss cases, the average PSNR must be calculated as follows:
\begin{equation}
{PSNR_{avg}} = 10 \times{\log _{10}}\left( {\frac{{{d^2}}}{{\mu_{MSE}}}} \right)\
\label{moneq6}
\end{equation}
where ${\mu_{MSE}}$ is the average of the frame MSE values.


\begin{itemize}
	\item The sample standard deviation of the MSE, ${\sigma _{MSE}}$, is a simple evaluation criterion that we use to quantify the temporal variability impact of video on QoE. It can be defined as follows:
\end{itemize}

\begin{equation}
{\sigma _{MSE}} = \sqrt {\frac{1}{K-1}\sum\limits_{k = 0}^{K-1} {{{(MS{E_k} - {\mu_{MSE}})}^2}} } \
\label{moneq5}
\end{equation}

\begin{itemize}
	\item SSIM \unskip~\cite{Wang} has been developed to measure the reconstructed frame visual quality. The idea of SSIM is to measure the structure similarity between two frames, rather than a pixel-to-pixel difference as for example the previous PSNR. The underlying assumption is that the human eye is more sensitive to changes in the structure of the image. Furthermore, the SSIM calculation is based on a combination of 3 comparisons: luminance, contrast and structure. The result varies from 0 to 1, where SSIM=1 corresponds to a perfect similarity.
\end{itemize}

Table~\ref{table-wrap-7afe70dec01a7c70bf76894f5c7a3b32} summarizes the main parameters of the carried out simulations. The parameters established for the system are as follows: the probability for each description, ${P_{Descrip}}$, is fixed at 0 for the odd description (D1) and 0.6 for the even one (D2). In regards to the thresholds, they are set to 20 packets for $qt{h_{low}}$ and 45 for $qt{h_{high}}$ knowing that the maximum packet number in the interface queue (IFQ) is 50, thus taking up Lin et al.\unskip~\cite{259576:5809274} proposed values.

\begin{table}[!htbp]
	\caption{{Simulation parameters of the VANET scenario.} }
	\label{table-wrap-7afe70dec01a7c70bf76894f5c7a3b32}
	\def\arraystretch{1}
	\ignorespaces 
	\centering 
	\begin{tabulary}{\linewidth}{LL}
		\hline 
		Parameters & Value\\
		\hline 
		Number of vehicles &
		100\\
		Radio-propagation model &
		TwoRayGround\\
		Video play time &
		10s\\
		Maximum Transfer Unit (MTU) &
		1024\\
		Routing protocol &
		AODV\\
		Transport protocol &
		UDP\\
		Used metrics &
		Peak signal to noise ratio (PSNR)\mbox{}\protect\newline Structural similarity(SSIM) \mbox{}\protect\newline Number of lost packets\mbox{}\protect\newline
		Standard deviation of the MSE\\
		\hline 
	\end{tabulary}\par 
\end{table}

\section{Performance evaluation }

Several simulations have been carried out to validate the effectiveness of the proposed mechanisms. The displayed results are those of the BQmall sequence with the highest frame rate. However, similar results for the other sequences are obtained. We have defined two distinct scenarios:
\begin{itemize}
	\item A first scenario considered severe where only the video is transmitted. The goal here is to study the behavior of different mapping methods by filling queues,
	\item A second scenario, more realistic where the video is transmitted along with other streams. In this scenario, the vehicular environment is more favorable than the first with a lower vehicle speed and a smaller distance between vehicles.
\end{itemize}

\subsection{Buffer filling state analysis}

In this section, we analyze the queues filling for the different mechanisms. Figures~\ref{figure-edca}, \ref{figure-static} and \ref{figure-adap} show the filling state of the queues and the frame by frame video PSNR for the EDCA, static and proposed algorithms, respectively. For the three figures, the left Y axis represents the buffers filling status in packets number and the right Y axis the PSNR in dB. The X axis represent time in seconds.

The fact that the EDCA mechanism uses only one queue to transmit the video generates a drop in the queue as soon as it is filled. This packet loss has a direct impact on the quality of the video, as illustrated on Figure~\ref{figure-edca}. Also, the error concealment mechanism makes it possible to keep an acceptable PSNR value when the temporal changes are not important. That we explain by the copy frame technique and the high frame rate.

On the other hand, for the EDCA algorithm this will not change in comparison with a single description coding (SDC) transmission. Indeed, the fact that we apply an AI encoding means that each image is encoded independently of the rest of the sequence. This translates into exactly the same flow obtained between SDC and MDC encodings. While the other two algorithms use the MDC scheme specificity in the mapping decision.

\bgroup
\fixFloatSize{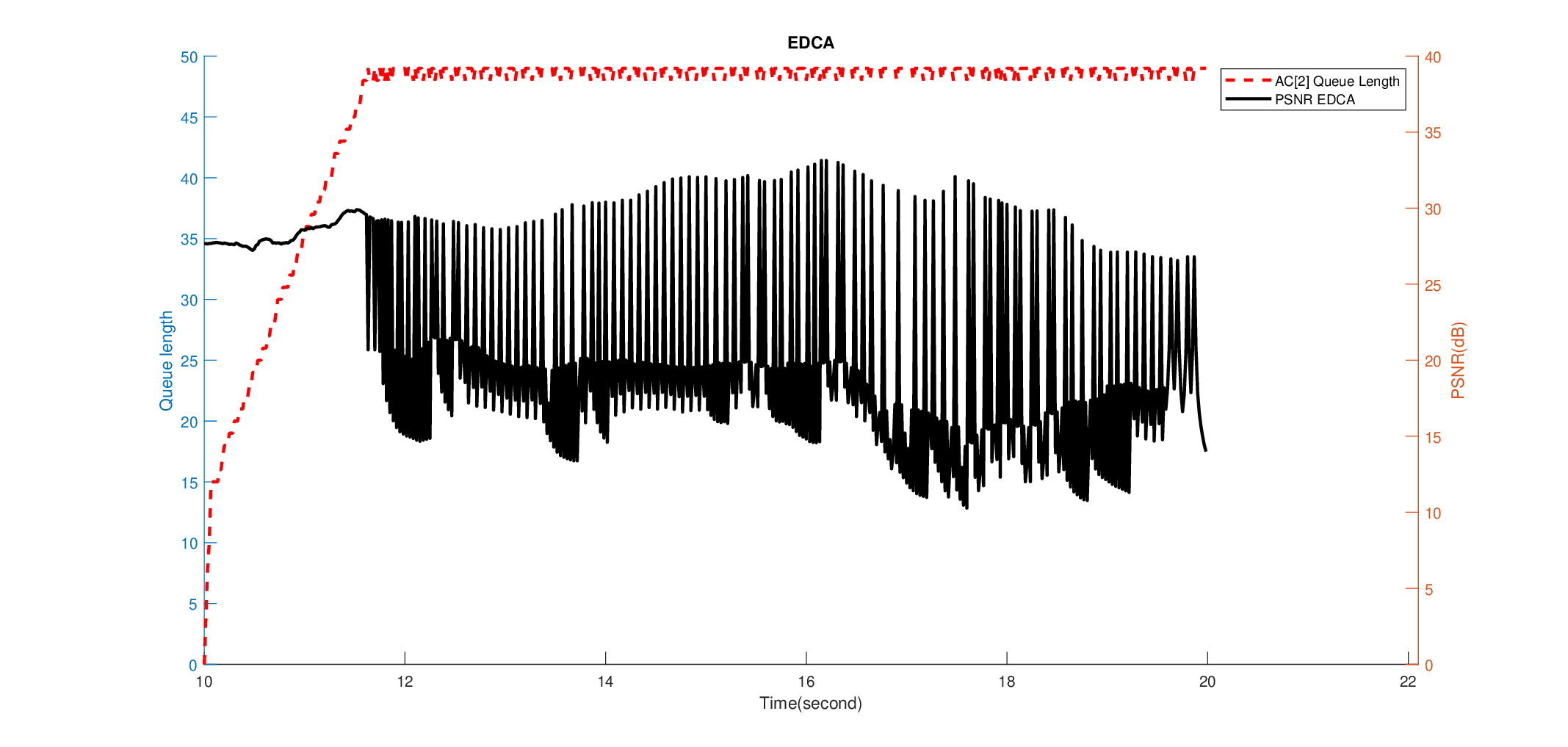}

\begin{figure*}[!htbp]
	\centering \makeatletter\IfFileExists{images/edca.eps}{\includegraphics[width=1.0\linewidth]{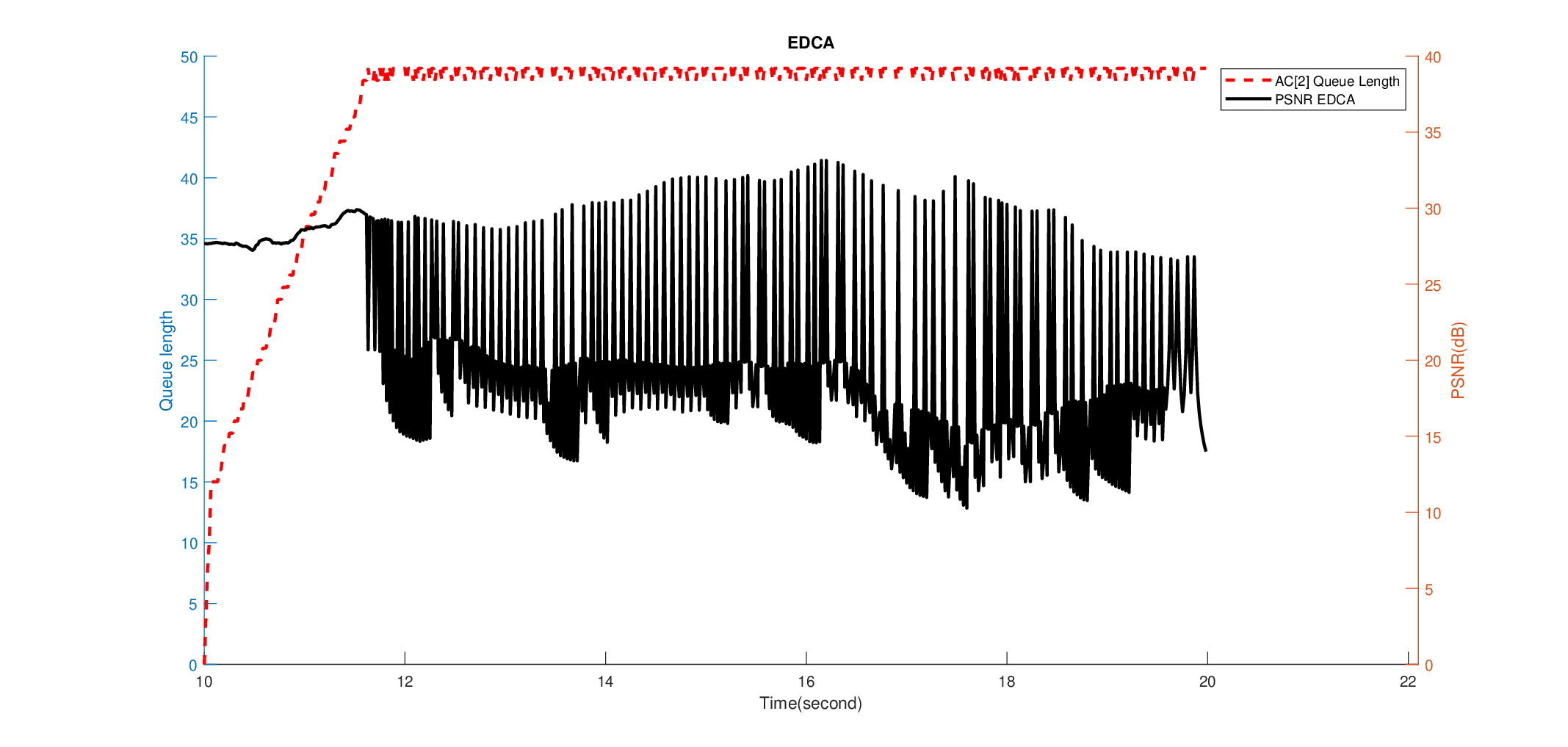}}{}
	\makeatother 
	\caption{{EDCA mapping algorithm: filling state of the queue and PSNR video quality.}}
	\label{figure-edca}
\end{figure*}
\egroup

The Figure~\ref{figure-static} allows us to see the improvement of the static method video PSNR. This is explained by the use of an additional AC compared to the EDCA method. We can nevertheless note the under-utilization of the AC with the highest priority:  \textit{AC}[2]. Also, the saturation of the  \textit{AC}[1] completely penalizes the disadvantaged video description. This mechanism allows the good reception of one of the two descriptions, which provide a video with 1/2 frame rate. However, it remains inefficient given the lost resources.

\bgroup
\fixFloatSize{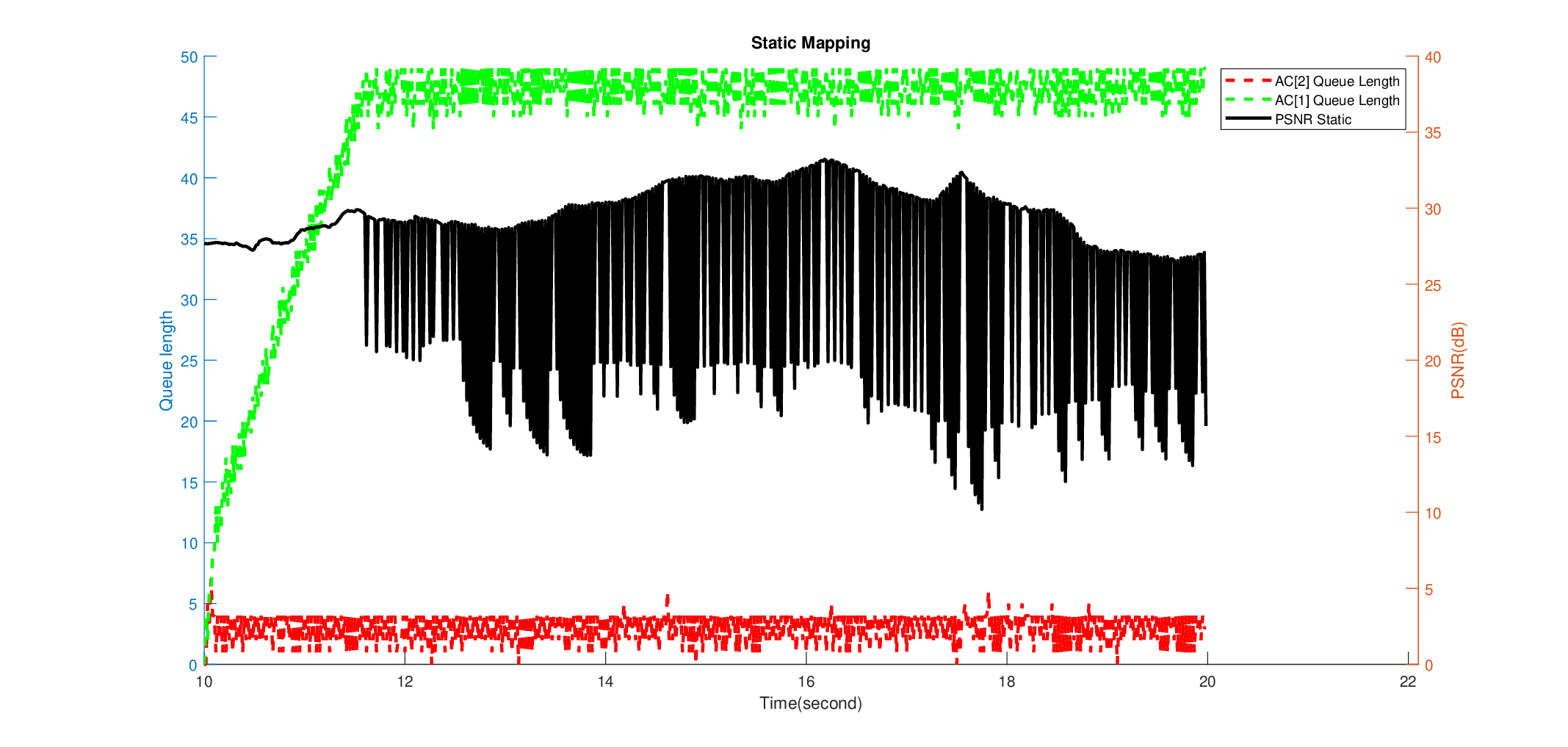}
\begin{figure*}[!htbp]
	\centering \makeatletter\IfFileExists{images/static.eps}{\includegraphics[width=1.0\linewidth]{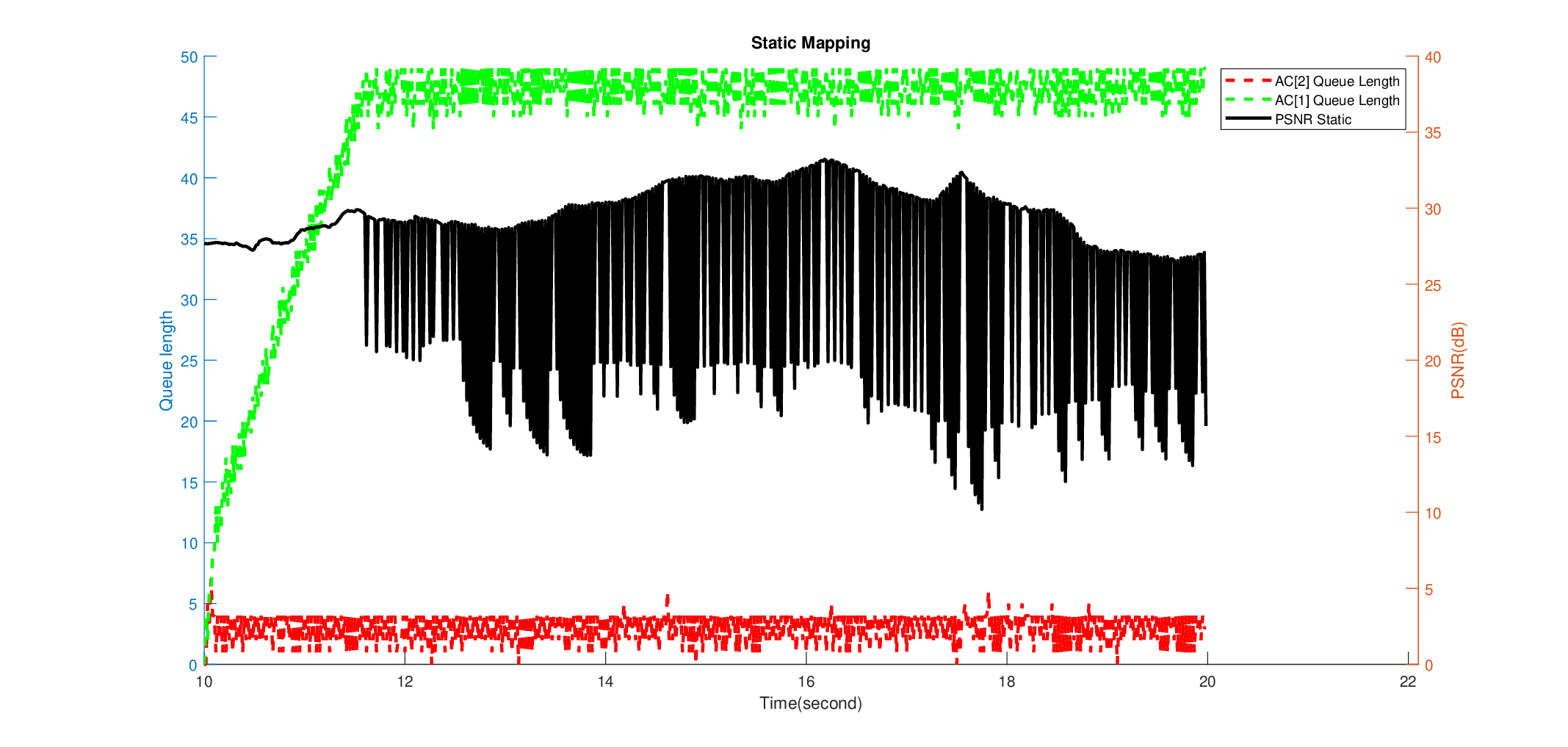}}{}
	\makeatother 
	\caption{{Static mapping algorithm: filling state of the queue and PSNR video quality.}}
	\label{figure-static}
\end{figure*}
\egroup

Figure~\ref{figure-adap} illustrates the behavior of the queues with the proposed adaptive mechanism, it makes full use of the resources made available by the multiple ACs and exploits the resources in priority order. This translates into better video quality than the other two systems. Switching from one queue to another allows to choose one description rather than the other.

\bgroup
\fixFloatSize{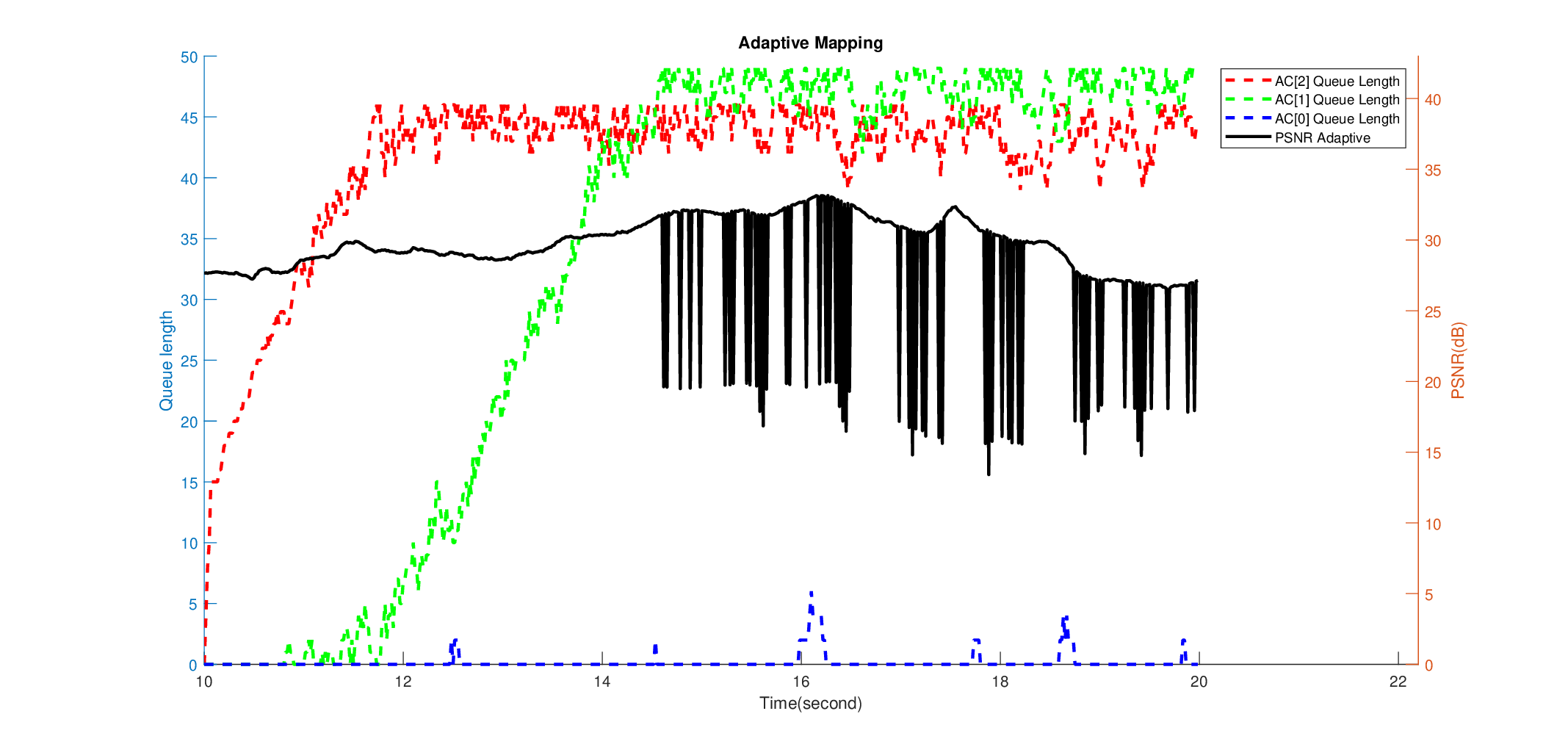}
\begin{figure*}[!htbp]
	\centering \makeatletter\IfFileExists{images/adap.eps}{\includegraphics[width=1.0\linewidth]{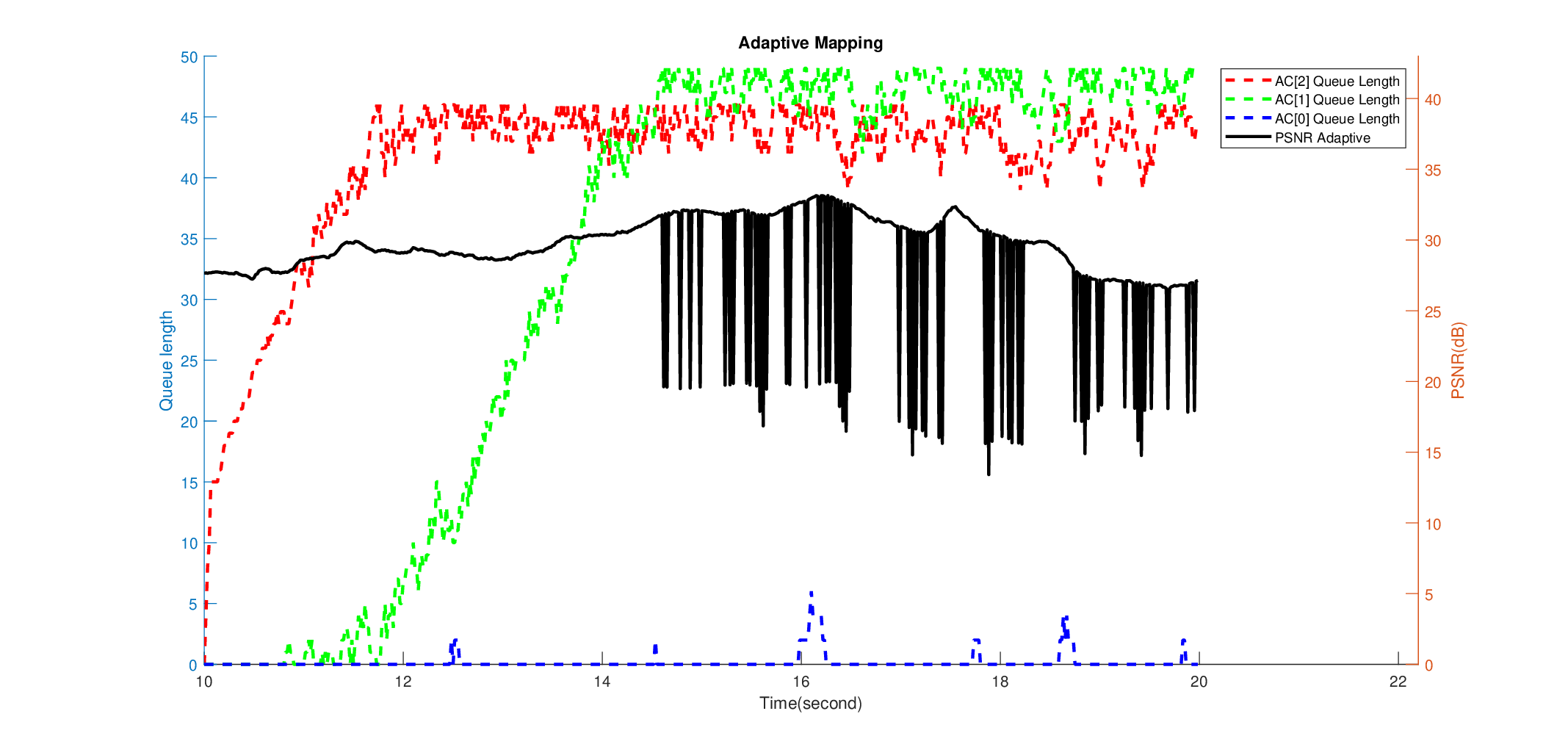}}{}
	\makeatother 
	\caption{{Proposed adaptive mapping algorithm: filling state of the queue and PSNR video quality.}}
	\label{figure-adap}
\end{figure*}
\egroup

Table~\ref{table-7} shows the standard deviation of the MSE, the average PSNR and the average SSIM for each mapping algorithm. We can see a PSNR gain of 7.75 dB between the EDCA and the proposed adaptive methods. Also, a gain of 0.2476 in SSIM. Finally, the difference in the standard deviation of the MSE illustrates the lower temporal variability of the proposed method that preserves the QoE. Indeed, we verify visually that, when considering our method, displayed video sequences are more fluid and exhibit less temporal artifacts like misalignment or frozen pictures.

\begin{table}[!htbp]
	\caption{{MSE standard deviation, average PSNR and average SSIM for each mapping algorithm (First scenario).} }
	\label{table-7}
	\def\arraystretch{1}
	\ignorespaces 
	\centering 
	
	\begin{tabular}{llll} 
		\hline
		Mapping Algorithm & ${\sigma _{MSE}}$ & Average SSIM & Average PSNR (dB)  \\ 
		\hline
		EDCA algorithm          & 436.28    & 0.7021     & 21.95              \\
		Static algorithm        & 302.07    & 0.8564     & 25.37              \\
		Proposed algorithm      & 125.66    &  0.9497        &  29.71              \\
		\hline
	\end{tabular}
\end{table}

\subsection{Multi-stream transmission}

In real networks, video usually coexists with other types of streams. For this, we simulate with video transmission, a voice traffic in the \textit{AC}[3] but also a TCP stream in the \textit{AC}[1] and a UDP stream on the \textit{AC}[0]. 

Figure~\ref{figure-41b2bcbae9c5aba4feae6a27060e6592} shows the frame-by-frame PSNR of the BQmall sequence transmitted with the three mapping methods. We notice a frequent degradation of the quality of the video transmitted with the EDCA which is due to the limited capacity of the video buffer. 

\bgroup
\fixFloatSize{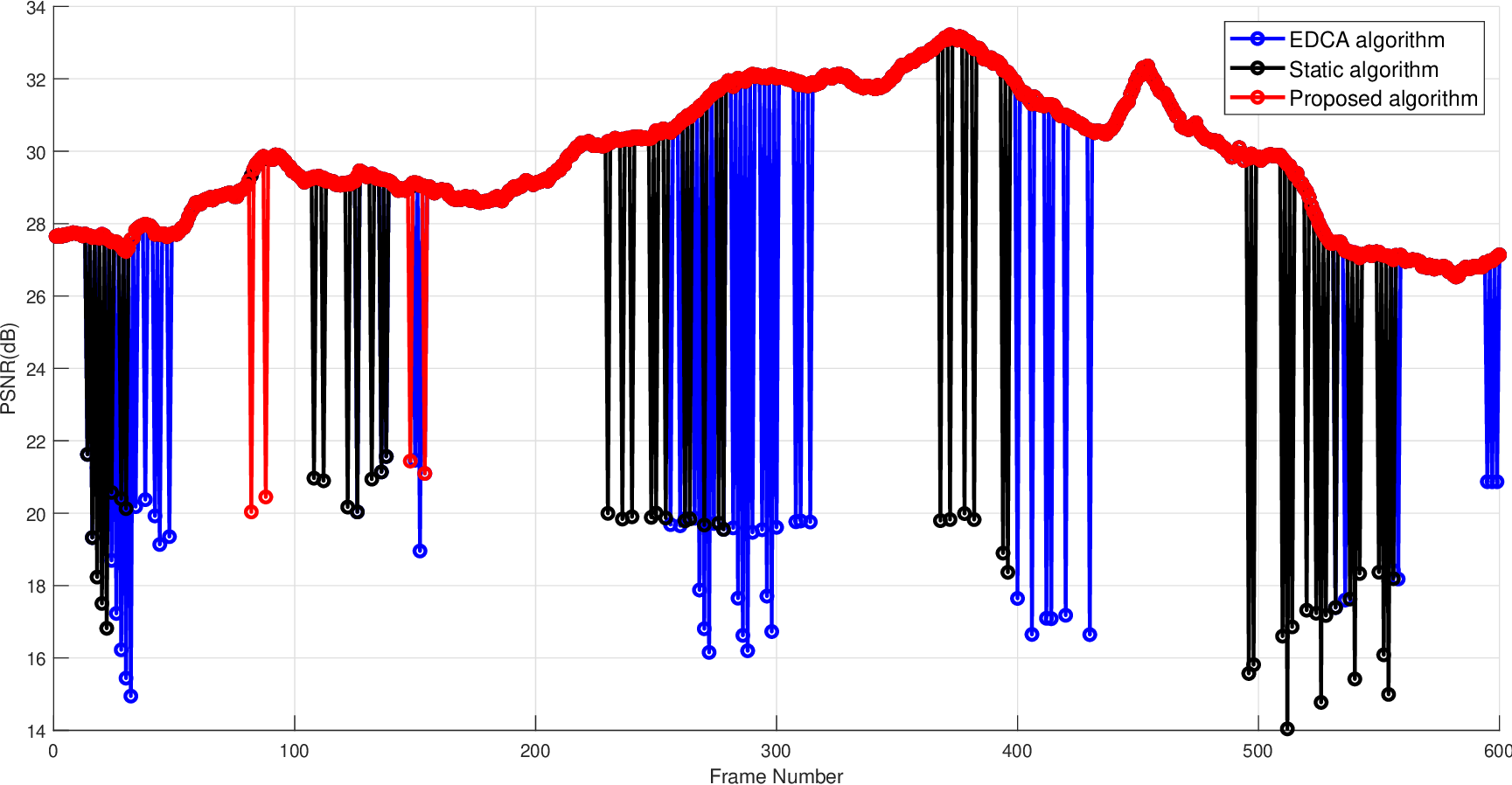}
\begin{figure}[!htbp]
	\centering \makeatletter\IfFileExists{images/47576080-277f-4564-9ac7-fdae86282e8b-upsnr_sequence_first_288_frames-0.eps}{\includegraphics[width=0.9\linewidth]{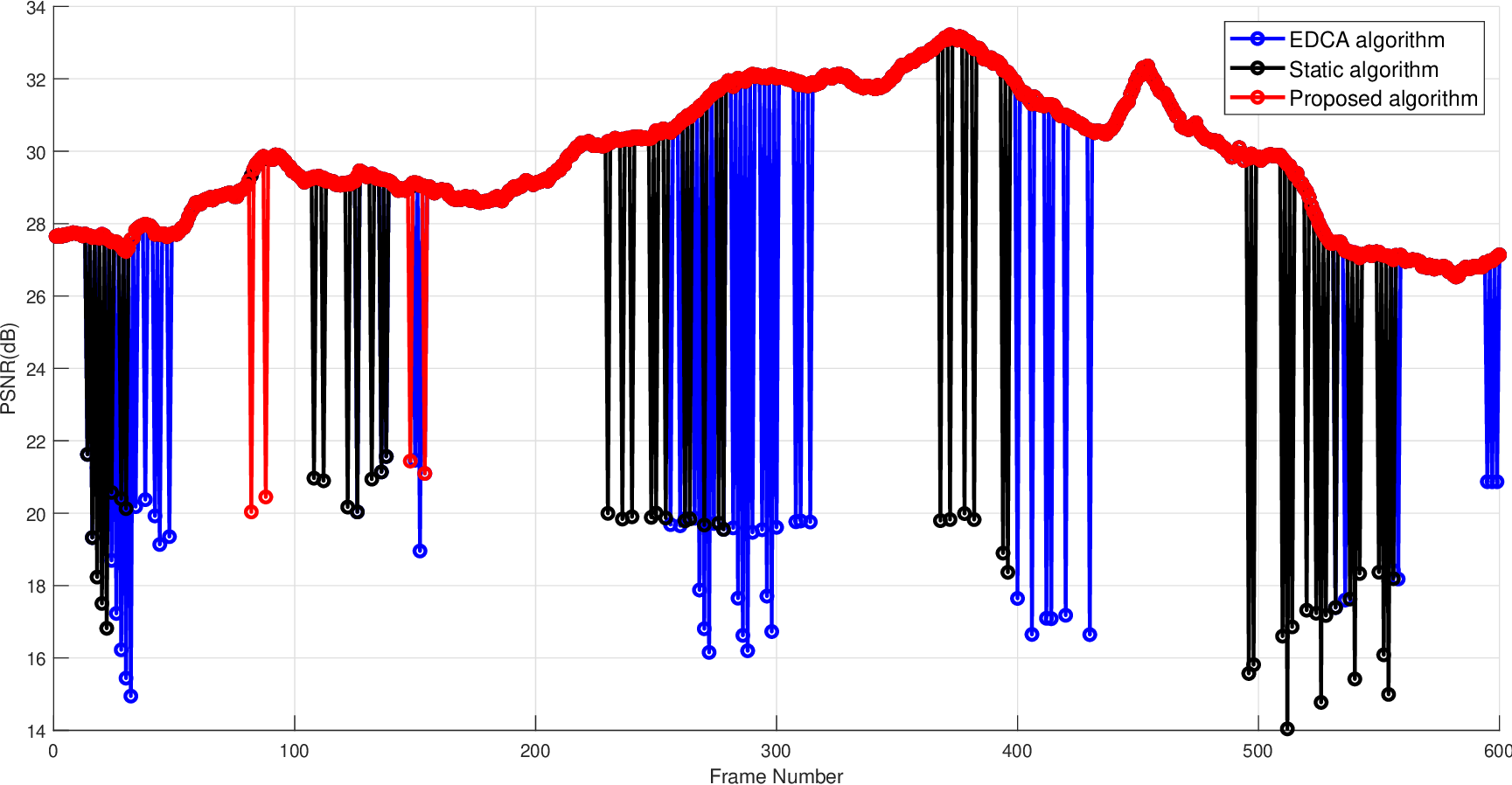}}{}
	\makeatother 
	\caption{{PSNR variation for the different mapping algorithm.}}
	\label{figure-41b2bcbae9c5aba4feae6a27060e6592}
\end{figure}
\egroup

For its part, the static method uses two queues \textit{AC}[2] and \textit{AC}[1] and this reduces the number of video packet lost especially for video packet of the favored description (D1). Nevertheless, the number of lost packets of the (D2) remains important. Indeed, the description transmitted in the \textit{AC}[1] has a larger number of lost packets. This is due: to the packets drop and to the delay limitation imposed on different video packet, which are dedicated to a low latency application. The fact that the \textit{AC}[1] takes more time to channel access brings a consequent delay to the video packet. Also, due to the under-exploitation of \textit{AC}[2] during the transmission of the video sequence, as explained previously. 
\mbox{}\protect\newline On the other hand, the proposed adapative method optimally utilizes the ACs of the 802.11p MAC and thus offers better quality than other systems. The copy frame that is set up at the receiving end reduces the loss packet effect. The results in Table~\ref{table-wrap-4aa459235b694b52d30c4aeb536fcfb7} show more the discrimination adopted in the packets mapping. Indeed, we can see the evaluation in terms of the number of lost packets and an evaluation of the video quality with the average SSIM and the average PSNR for the three mapping methods. 

\begin{table}[!htbp]
	\caption{{ MSE standard deviation, Average PSNR, Average SSIM and number of lost packets for each mapping algorithm (Second scenario)} }
	\label{table-wrap-4aa459235b694b52d30c4aeb536fcfb7}
	\def\arraystretch{1}
	\ignorespaces 
	\centering 
	\begin{tabulary}{\linewidth}{LLLLLLLL}
		\hline 
		\multirow{2}{*}{\begin{tabular}[c]{@{}l@{}}Mapping \\ Algorithm\end{tabular}} &
		\multirow{2}{*}{\begin{tabular}[c]{@{}l@{}}${\sigma _{MSE}}$\end{tabular}} &
		\multirow{2}{*}{\begin{tabular}[c]{@{}l@{}}Average \\SSIM \end{tabular}} & \multirow{2}{*}{\begin{tabular}[c]{@{}l@{}}Average\\ PSNR (dB) \end{tabular}} & \multirow{2}{*}{\begin{tabular}[c]{@{}l@{}} Transmitted \\ packets \end{tabular}} & \multicolumn{3}{l}{Lost packets} \\
		&                   &                   &                        & &D1  & D2 & Tl \\
		\hline 
		EDCA algorithm &
		143.81 &
		0.9439 &
		28.92 &
		&
		249 &
		273 &
		522\\
		Static algorithm &
		104.63 &
		0.9511 &
		30.37 &
		3600 &
		0 &
		95 &
		95\\
		Proposed algorithm &
		24.41 &
		0.9686 &
		32.64 &
		&
		2 &
		9 &
		11\\
		\hline 
	\end{tabulary}\par 
\end{table}

We notice a significant gain in the number of packets correctly received with the static and proposed algorithms. Indeed, the static method is able to better preserve the most important description but remains less effective than the proposed adaptive method in terms of correctly received packets and video quality. This has a direct impact on video quality as shown by the PSNR, SSIM and MSE standard deviation.

As previously explained the different simulations were performed with different video sequences, on average the gain brought by the proposed method in comparison to the EDCA method is of the order of 2.34 dB for the chosen transmission scheme. We can establish that the proposed adaptive method is even more efficiency when the transmission channel is more hostile. Visually, we verify that the quality of the decoded sequences fluctuates much less. This results in an enhanced quality of experience for the final viewer.

\section{Conclusion}

In this work, we propose the use of the MDC in low latency application over vehicular networks; to do this we presented an algorithm that allow the improvement of the transmission. Indeed, the proposed adaptive algorithms allow cross-layer packet classification based on the IEEE 802.11p protocol. The proposed strategy is based on the mapping of video packets in the most suitable ACs in order to propose a better QoS. The enhancement we propose uses application layer information at the MAC layer in a cross-layer scheme. Indeed, the information on the video description and the MAC AC buffer filling allow the proposed algorithm to choose the best option for video mapping. The results established in a realistic vehicular environment show an improvement of the QoS and also an improvement of the video quality. As a perspective, it would be interesting to investigate the efficiency of more complex MDC schemes with video encoding structures based on temporal video prediction. It would also be interesting to continue to use the differentiation of the MDC in the routing protocol with routes adapted to each of the descriptions. This can be done in a three-tier system, in addition to the current two-tier one.

\section*{References}

\bibliographystyle{vancouver}

\bibliography{article}

\end{document}